\documentclass{article}

\usepackage{arxiv}

\usepackage[utf8]{inputenc} % allow utf-8 input
\usepackage[T1]{fontenc}    % use 8-bit T1 fonts
\usepackage{hyperref}       % hyperlinks
\usepackage{url}            % simple URL typesetting
\usepackage{booktabs}       % professional-quality tables
\usepackage{amsfonts}       % blackboard math symbols
\usepackage{nicefrac}       % compact symbols for 1/2, etc.
\usepackage{microtype}      % microtypography
\usepackage{lipsum}
\usepackage{graphicx}
\graphicspath{ {./images/} }
\usepackage{amsmath}
\usepackage{amssymb}
\usepackage{multirow}
\usepackage{booktabs} 
\usepackage{tabularx}
\usepackage{adjustbox}
\usepackage{float}
\usepackage{svg}
\usepackage{graphicx,verbatim}
\usepackage{soul}
\usepackage{ragged2e}
\usepackage[table]{xcolor}
\usepackage{soul}
\definecolor{oraclegray}{RGB}{230,230,230} % light gray
\usepackage{enumitem}
\setlist{nosep, leftmargin=14pt}
\usepackage{svg}
\usepackage{comment}
\usepackage{bm}
\usepackage{mwe} % to get dummy images

\usepackage{pifont} % add this in the preamble for check/cross

\newcommand{\cmark}{\ding{51}}%
\newcommand{\xmark}{\ding{55}}%

\title{Weakly Supervised Concept Learning with Class-Level Priors for Interpretable Medical Diagnosis}

\author{
 Md Nahiduzzaman \\
  School of Computing Technologies\\
  RMIT University\\
  Melbourne VIC 3000, Australia \\
  \texttt{s4045807@student.rmit.edu.au} \\
  \And
 Steven Korevaar \\
  School of Computing Technologies\\
  RMIT University\\
  Melbourne VIC 3000, Australia \\
  \texttt{steven.korevaar@rmit.edu.au} \\
  \And
  Alireza Bab-Hadiashar \\
  School of Engineering\\
  RMIT University\\
  Melbourne VIC 3000, Australia \\
  \texttt{alireza.bab-hadiashar@rmit.edu.au} \\
     \And
 Ruwan Tennakoon \\
  School of Computing Technologies\\
  RMIT University\\
  Melbourne VIC 3000, Australia \\
  \texttt{ruwan.tennakoon@rmit.edu.au} \\
}

\begin{document}
\maketitle
\begin{abstract}
Human-interpretable predictions are essential for deploying AI in medical imaging, yet most interpretable-by-design (IBD) frameworks, require concept annotations for training data, which are costly and impractical to obtain in clinical contexts. Recent attempts to bypass annotation, like zero-shot vision–language models or concept-generation frameworks, struggle to capture domain-specific medical features, leading to poor reliability. In this paper, we propose a novel \emph{Prior-guided Concept Predictor} (PCP), a weakly supervised framework that enables concept answer prediction without explicit supervision or reliance on language models. PCP leverages class-level concept priors as weak supervision and incorporates a refinement mechanism with KL divergence and entropy regularization to align predictions with clinical reasoning. Experiments on PH2 (dermoscopy) and WBCatt (hematology) show that PCP improves concept-level F1 by over 33\% compared to zero-shot baselines, while delivering competitive classification performance on four medical datasets (PH2, WBCatt, HAM10000, and CXR4) relative to fully supervised CBMs and V-IP. 
\end{abstract}

% keywords can be removed

\section{Introduction}
\label{sec:intro}

Deep learning (DL) has achieved remarkable success in medical imaging, yet most models operate as ``black boxes'', limiting clinical trust and adoption. Interpretable-by-design (IBD) models, such as \emph{Concept Bottleneck Models (CBM)}~\cite{koh2020concept} and \emph{Variational-Information Pursuit (V-IP)}~\cite{chattopadhyayvariational}, address this by mapping raw image features to human-understandable concepts (e.g., \emph{blue-whitish veil}, \emph{irregular streaks}) and using them for classification. CBMs require all concepts to be predicted before making a decision, while V-IP adaptively queries only the most informative ones. Both approaches provide meaningful explanations aligned with clinical reasoning. Building on these ideas, the Multimodal Interpretable Concept Alignment (MICA)~\cite{bie2024mica} and Concept-Based Vision–Language Model (CBVLM)~\cite{patricio2025cbvlm} require few-shot or partially annotated concepts. 
However, collecting per-concept annotations at scale remains difficult in medical imaging, as concepts are subtle (there is often disagreement even amongst experts) and experts have limited time and resources. Thus, the central challenge is achieving concept prediction without explicit concept supervision. 

Recent works attempt to relax this dependency. 
Post-hoc CBMs~\cite{yuksekgonulpost} use Concept Activation Vectors from external datasets but fail when no curated concept bank exists. 
Language-guided models such as LaBo-CBM~\cite{yang2023language} use large language models (LLMs) to define textual concepts, while V2C-CBM~\cite{he2025v2c} employs VLMs to tokenize visual concepts. Similarly, ConceptCLIP~\cite{nie2025conceptclip} learns image--text--concept embeddings via medical ontologies, but its reliance on predefined vocabularies and alignment quality limits generalization to subtle clinical findings. 
Recent V-IP extensions, including Bootstrapped~\cite{chattopadhyay2024bootstrapping} and Learned-Query V-IP~\cite{kolek2025learning}, integrate VLMs like CLIP to answer or learn interpretable queries but still rely on textual supervision. 
In a different line of work, the Visual Perception Module (VPM)~\cite{yu2024zero} introduces attention-based refinement for concept-guided zero-shot classification but assumes access to \emph{ground-truth concept annotations}, differing from our goal of predicting concept answers without annotations or VLM reliance.

\begin{figure}[ht]
    \centering
    \includegraphics[width=\linewidth]{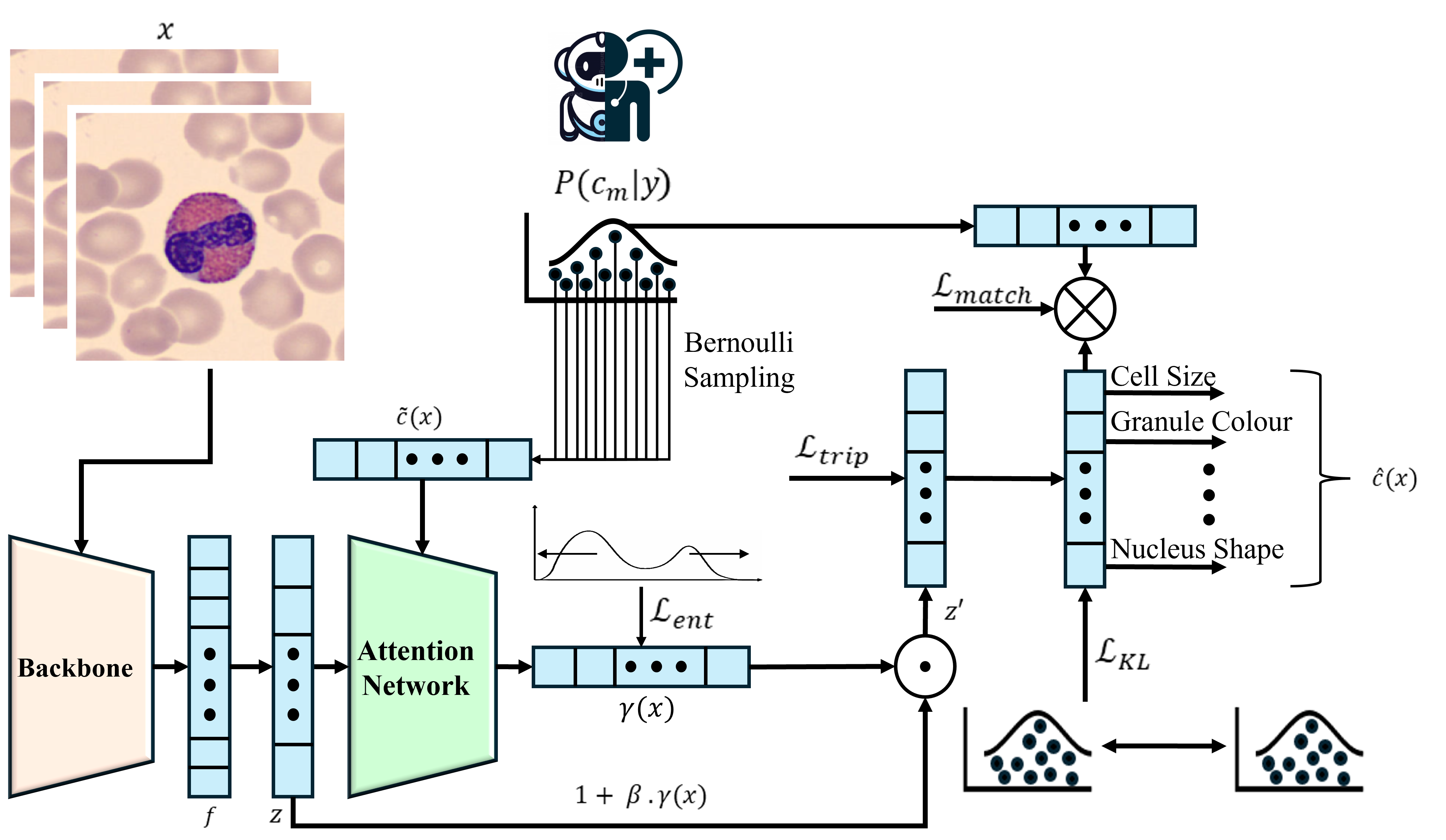}
    \caption{Overview of the proposed framework. 
Given an input image, a ResNet backbone extracts features, which are projected into a concept space and refined through a prior-guided attention mechanism. 
% Surrogate concept vectors, sampled from class-level priors, serve as probabilistic guides that modulate the attention weights. 
% The attention weights are modulated probabilistically based on class-level priors.
Attention is computed by combining projected features with probabilistic vectors sampled from concept-priors and subsequently refined (Eq.~\ref{eq:residual-refinement}). 
KL regularization aligns predicted concepts with priors, while entropy minimization sharpens attention distributions. 
% At inference, only the image is required, and the model directly outputs concept probabilities.
}
    \label{fig:framework}
\end{figure}

\textit{In this work, we address the question: Can medical concept prediction be achieved without explicit concept supervision or reliance on VLM guidance?} 
We propose a novel weakly supervised \textit{Prior-guided Concept Predictor (PCP)} framework (Fig.~\ref{fig:framework}) for medical concept prediction \emph{without concept-level annotations or vision--language supervision}. 
PCP learns concepts from image features using class-level concept priors, aligning predicted concept distributions with known statistics while preserving image-specific variability. These priors can be derived from domain experts, dataset–level statistics, or automated knowledge sources, and are far easier to obtain (e.g., from a single expert consultation) than exhaustive concept-level annotations, thus bypassing the prohibitive cost of manual supervision.
Two regularizers, a KL-divergence loss and an entropy loss, promote alignment and concept selectivity.
In contrast to VPM, which focuses only on the most discriminative features, our proposed PCP framework predicts \emph{all clinically relevant concepts}, preserving interpretability.

We validate PCP on four medical datasets spanning different imaging modalities:
PH2 (dermoscopy) and WBCatt (hematology), which include ground-truth concept annotations for evaluating concept-level prediction accuracy,
and HAM10000 (dermoscopy) and CXR4 (chest X-ray), which lack concept labels and are evaluated using LLM-derived class-level priors.
Across all datasets, PCP outperforms zero-shot VLM baselines (CLIP, SigLIP, BioMedCLIP, ConceptCLIP) in concept prediction 
and achieves classification performance comparable to fully supervised CBM and V-IP models, 
demonstrating that reliable and interpretable concept reasoning is possible even without explicit concept supervision.

In summary, the major contribution of this work is the introduction of the \textit{PCP}, a weakly supervised framework that enables medical concept prediction without concept-level annotations or VLM supervision by aligning image-predicted concepts with class-level priors through a composite loss, achieving interpretable and competitive performance across multiple medical imaging datasets.

\section{Methodology}
\subsection{Problem Setup}
\label{sec:PS}

The goal of this research is to predict an outcome label from an input image through a set of human–interpretable concepts.  
Given a training dataset
\(
\mathcal{D}_{\textrm{train}} = \{(x_i,y_i)\}_{i=1}^N,
\)
where each image $x_i \in \mathcal{X}$ is paired with an outcome label $y_i \in \mathcal{Y}$, and the label space has cardinality $|\mathcal{Y}| = L$, we assume the existence of $M$ human–interpretable concepts, denoted by  
$\mathcal{C} = \{c_1, c_2, \ldots, c_M\}$. 

%each representing a clinically meaningful visual attribute  (e.g., ``blue–whitish veil'' or ``irregular streaks'' in dermoscopy).
A \textit{concept predictor} maps an input image $x$ to a probability vector of concept activations such that:
\[
\hat{\mathbf{c}}(x) = 
[\hat{c}_1(x), \ldots, \hat{c}_M(x)] \in [0,1]^M,
\quad \hat{\mathbf{c}} : \mathcal{X} \rightarrow [0,1]^M.
\]
The predicted concept vector is then used for classification.
% \[
% \mathbf{clf} : [0,1]^M \rightarrow \mathcal{Y},
% \]
% Traditional CBMs or V-IP~\cite{koh2020concept,chattopadhyayvariational} rely on supervised learning to train a concept predictor that outputs $\hat{\mathbf{c}}(x)$ using ground-truth concept labels. However, such labels are rarely available in medical imaging due to the high cost and uncertainty of annotation.
%Moreover, weak supervision from vision–language models often fails to capture subtle clinical attributes~\cite{yuksekgonulpost} (see Table~\ref{tab:concept-results}).

% This is an ill-posed problem that cannot be solved without introducing additional constraints. 
To train our model without sample-wise concept labels, we rely on concept-wise priors, which represent the likelihood of each concept appearing in an image belonging to a specific class.
%we use \emph{concept-wise priors} that are the likelihood of each concept appearing in an image of a specific class. 
For each class $y \in \mathcal{Y}$, the probability of a concept $c_m \in \mathcal{C}$ appearing is given by: $P(c_m \mid y) \in [0,1]$.

\subsection{Proposed Model}
\label{sec:model}
 % Our goal is to predict whether each human-interpretable concept is present or not in an input image, without requiring ground-truth concept annotations. 
Our concept prediction model combines a backbone encoder (a projection into concept space) and a prior-guided refinement mechanism.  
We employ a ResNet backbone pretrained on ImageNet to extract visual features from the input image 
\( x \in \mathbb{R}^{C \times H \times W} \).
The classification head of the backbone is replaced with two bias-free linear layers to reduce dimensionality and map the features onto concept space, the result being $\mathbf{z}$.

We use Bernoulli sampling from class-specific priors to generate surrogate concept vectors. For an image with label $y$, the surrogate vector is defined as:
\begin{equation}
\tilde{\mathbf{c}}(x) \sim \text{Bernoulli}(P(c_m \mid y)), 
\quad \tilde{\mathbf{c}}(x) \in \{0,1\}^M,
\label{eq:surrogate}
\end{equation}
These sampled vectors act as probabilistic substitutes for the unobserved ground-truth concept annotations, acknowledging that not all concepts associated with a class appear in every instance.

We use element-wise multiplication to combine the projected image features with the surrogate concept vectors, akin to VPM~\cite{yu2024zero}:
\begin{equation}
\gamma(x) = \text{softmax}(\mathbf{z} \odot \tilde{\mathbf{c}}(x)), 
\quad \gamma(x) \in \mathbb{R}^M,
\label{eq:attention}
\end{equation}
% where $\odot$ denotes element-wise multiplication. 
The attention vector $\gamma(x)$ highlights the concepts most relevant for predicting class $y$.
Directly masking with $\gamma(x)$ as in \cite{yu2024zero} risks discarding weak but potentially informative concepts. To mitigate this, we adopt a residual refinement mechanism:
\begin{equation}
\mathbf{z}' = \mathbf{z} \odot \big(1 + \beta \cdot \gamma(x)\big),
\quad \mathbf{z}' \in \mathbb{R}^M,
\label{eq:residual-refinement}
\end{equation}
where $\beta$ is a tunable refinement strength parameter selected on the validation set for each dataset, amplifying important concepts while avoiding complete suppression of weaker but potentially informative ones, thereby improving stability and robustness.

Finally, the refined features are passed through a linear predictor with sigmoid activation:
\begin{equation}
\hat{\mathbf{c}}(x) = \sigma(W_c \mathbf{z}'), 
\quad \hat{\mathbf{c}}(x) \in [0,1]^M,
\label{eq:concept-prediction}
\end{equation}
where each entry $\hat{c}_m(x)$ estimates the probability that concept $c_m$ is present in image $x$. During training, surrogate vectors provide weak supervision but during inference, no priors are required. The model takes only an image as input and directly outputs $\hat{\mathbf{c}}(x)$, providing concept-level predictions without requiring concept labels.

\subsection{Training Objective}
\label{sec:training}

% Based on the refined feature vectors $\mathbf{z}'$ and predicted concepts $\hat{c}(x)$, 
We define a composite training loss combining Triplet, Class-Matching, KL, and Entropy terms to jointly promote discriminative embeddings, 
concept predictions consistent with priors, and regularized attention distributions.  
% Our objective does not directly target class prediction; rather, 
% it guides the model to learn clinically meaningful concepts, where some concepts 
% are typically present and others are typically absent in each class.  

\textbf{(1) Triplet Loss.} Following~\cite{yu2024zero}, we apply a triplet loss on the refined embeddings $\mathbf{z}'$. 
For each anchor $\mathbf{z}'_i$, we randomly select a positive example 
$\mathbf{z}'_j$ from the same class and a negative example $\mathbf{z}'_m$ 
from a different class. The triplet loss is defined as
\begin{equation}
\mathcal{L}_{\text{trip}} = \frac{1}{N} \sum_{i=1}^N 
\big[ \|\mathbf{z}'_i - \mathbf{z}'_j\|_2 - \|\mathbf{z}'_i - \mathbf{z}'_m\|_2 + \mu \big]_+,
\label{eq:triplet}
\end{equation}
where $\|\cdot\|_2$ denotes the Euclidean norm, $\mu$ is the margin 
hyperparameter, and $[x]_+ = \max(x,0)$.  
This ensures that concept-driven embeddings from the same class are pulled 
closer together while those from different classes are pushed apart.

\textbf{(2) Class Matching Loss.} A class matching loss is computed via a dot product between predicted concept vectors and class-priors. This results in, $\mathbf{s}_i = [s_{i,1}, \ldots, s_{i,L}]$, a similarity vector for image $i$ across all $L$ classes.  
\begin{equation}
s_{i,k} = \left< \hat{\mathbf{c}}(x_i),  P(c_m|k) \right>,
\quad \mathbf{s}_i \in \mathbb{R}^L,
\label{eq:similarity}
\end{equation}
We then convert these logits into a probability distribution with softmax and apply cross-entropy:
\begin{equation}
\mathcal{L}_{\text{match}} \;=\; 
\frac{1}{N} \sum_{i=1}^N \mathrm{CE}\!\big(\,\mathrm{softmax}(\mathbf{s}_i),\, y_i \big).
\label{eq:class-loss}
\end{equation}
Thus, $s_{i,k}$ measures how well the predicted concept vector of image $i$ matches the prior signature of class $k$.

\textbf{(3) KL Regularization.} We use KL Regularization on the predicted concept probabilities to match the expected class-priors
by minimizing the KL divergence between the \emph{class priors} $P(c_k \mid y)$ and the batch-wise
\emph{empirical mean predictions} $\bar{P}(\hat{c}_k \mid y)$ over samples of class $y$:
\begin{align}
\mathcal{L}_{\text{KL}}
&=
\frac{1}{|\mathcal{Y}_{\text{batch}}|}
\sum_{y \in \mathcal{Y}_{\text{batch}}}
\Bigg[
\sum_{g \in \mathcal{G}}
\sum_{k \in g}
P(c_k \mid y)
\log\tfrac{P(c_k \mid y)}{\bar{P}(\hat{c}_k \mid y)}
\Bigg],
\label{eq:kl-loss}
\end{align}
where $\mathcal{G}$ denotes the set of mutually exclusive concept groups 
(e.g., \emph{cell size}=\{big, small\},
\emph{nucleus shape}=\{irregular, bilobed, multilobed, band, indented, round\}).

\textbf{(4) Entropy Loss.}  
Finally, we penalize the entropy of the attention distribution 
$\gamma(x)$ to encourage sharper weighting across concepts:
\begin{equation}
\mathcal{L}_{\text{ent}} = -\frac{1}{N}\sum_{i=1}^N \sum_{m=1}^M 
\gamma_{i,m}(x)\,\log \gamma_{i,m}(x).
\label{eq:entropy}
\end{equation}
This pushes irrelevant concepts toward $0$ while amplifying relevant 
ones toward one.

The overall objective combines these four components:
\begin{equation}
\mathcal{L} = \mathcal{L}_{\text{trip}} + \mathcal{L}_{\text{match}} + 
\lambda_{KL} \mathcal{L}_{\text{KL}} + \lambda_{\text{ent}} \mathcal{L}_{\text{ent}},
\label{eq:final-loss}
\end{equation}
where $\lambda_{KL}$ and $\lambda_{\text{ent}}$ balance the contribution 
of KL regularization and entropy minimization. 
\section{Experiments}
\label{sec:experiments}

\noindent
We evaluate the proposed framework on four medical imaging datasets: 
PH2~\cite{mendoncca2013ph}, a small binary dermoscopy dataset with 200 images and 8 clinical concepts; 
WBCatt~\cite{tsutsui2023wbcatt}, a large-scale hematology dataset with 10,198 images from five white blood cell types and 29 morphological concepts; 
HAM10000~\cite{tschandl2018ham10000}, a dermoscopic dataset with 7,818 images from two diagnostic classes (nevus and melanoma) and 25 concepts; and 
CXR4~\cite{T.rahman2020reliable,cohen2020covidProspective}, a chest X-ray dataset with 7,135 images from four diagnostic classes (Normal, COVID-19, Pneumonia, and Tuberculosis) and 29 radiological concepts. 
For the HAM10000 and CXR4 datasets, concept priors were generated using an LLM (e.g., ChatGPT) based on clinically meaningful class descriptions.

We report the F1-score for classification performance, and accuracy (Acc) and F1-score for concept prediction performance. 
All models are trained for 200 epochs using the Adam optimizer with a learning rate of \(5\times10^{-4}\). 
The entropy regularization weight is set to \(\lambda_{\text{ent}} = 0.01\) for all datasets except WBCatt (\(0.03\)), and the KL divergence weight is fixed at \(\lambda_{\text{KL}} = 0.3\) for all datasets (tuned on validation set). 
We use ResNet-101 as the backbone for PH2, HAM10000, and CXR4, and ResNet-34 for WBCatt. 

\subsection{Results and Discussion}
\label{sec:results}

All experiments were conducted with three random seeds (except for CLIP-based models, which are deterministic) and train-validation-test splits to ensure robustness. 
We evaluate both concept prediction and downstream classification performance on WBCatt and PH2, where concept-level annotations are available, to verify that our method produces interpretable and predictive concept representations.  
In our setup, Baseline refers to models trained with ground-truth concepts; Vanilla-CBM/V-IP denote the standard CBM~\cite{koh2020concept} and V-IP~\cite{chattopadhyayvariational} trained with full concept supervision; and PCP-CBM and PCP-V-IP correspond to our proposed weakly supervised models trained only with class-level priors.

\begin{table}[h]
\centering
\caption{Concept prediction performance on PH2 and WBCatt. 
Best results are in \textbf{bold}.}
\label{tab:concept-results}
\scriptsize
\resizebox{\linewidth}{!}{
\begin{tabular}{lcccccc}
\toprule
\textbf{Model} & $\mathcal{L}_{\text{KL}}$ & $\mathcal{L}_{\text{ent}}$ & \multicolumn{2}{c}{\textbf{WBCatt}} & \multicolumn{2}{c}{\textbf{PH2}} \\
\cmidrule(lr){4-5} \cmidrule(lr){6-7}
 &  &  & Acc(\%) & F1(\%) & Acc(\%) & F1(\%) \\
\midrule
\rowcolor{oraclegray}
Baseline & -- & -- & $95.29_{\pm 1.02}$ & $88.77_{\pm 0.95}$ & $84.38_{\pm 1.21}$ & $75.38_{\pm 0.98}$ \\
CLIP~\cite{radford2021learning} & -- & -- & $41.25$ & $41.41$ & $56.56$ & $32.96$ \\
SigLIP~\cite{zhai2023sigmoid} & -- & -- & $58.05$ & $4.58$ & $61.25$ & $24.79$ \\
BioMedCLIP~\cite{zhang2023biomedclip} & -- & -- & $55.52$ & $12.21$ & $61.56$ & $43.81$ \\
ConceptCLIP~\cite{nie2025conceptclip} & -- & -- & $48.43$ & $26.34$ & $58.75$ & $39.33$  \\
\midrule
\multirow{4}{*}{PCP (Ours)} 
 & \xmark & \xmark & $84.57_{\pm 1.42}$ & $66.68_{\pm 1.21}$ & $71.31_{\pm 1.86}$ & $52.71_{\pm 1.57}$ \\
 & \xmark & \cmark & $81.60_{\pm 1.11}$ & $57.68_{\pm 1.02}$ & $70.44_{\pm 1.37}$ & $47.23_{\pm 0.73}$ \\
 & \cmark & \xmark & $89.21_{\pm 1.32}$ & $77.01_{\pm 0.67}$ & $72.38_{\pm 1.13}$ & $65.41_{\pm 1.52}$ \\
 & \cmark & \cmark & \bm{$90.24_{\pm 0.38}$} & \bm{$79.00_{\pm 0.96}$} & \bm{$74.46_{\pm 1.74}$} & \bm{$69.02_{\pm 1.31}$} \\
\bottomrule
\end{tabular}}
\end{table}

\noindent \textbf{Concept Prediction.}  
% Our framework explicitly predicts \emph{concept answers} (present/absent), unlike models such as LaBo-CBM~\cite{yang2023language}, or ProbCBM~\cite{yuksekgonulpost}, which rely on LLMs to generate concepts and VLMs to obtain zero-shot answers, therefore 
In Table~\ref{tab:concept-results} we compare against zero-shot VLM-based methods (CLIP~\cite{radford2021learning}, SigLip \cite{zhai2023sigmoid}, BioMedCLIP~\cite{zhang2023biomedclip}, ConceptCLIP~\cite{nie2025conceptclip}). As is seen, these methods perform poorly in medical imaging, likely
%due to weak discriminative image–text alignment , inability to represent the absence of features, 
caused by a lack of medical data seen during pretraining. %subtle clinical attributes found in medical imagery. 
However, by learning from minimal extra data in the from of concept-class priors, our prior-guided PCP framework yields stronger and more clinically consistent predictions with concept-level accuracy improvements of more than 30\% compared to zero-
shot baselines. 
\\
\textbf{Ablation Study.} 
The ablation study highlights the complementary role of both regularizers. Removing $\mathcal{L}_{\text{KL}}$ causes predicted concept distributions to deviate from class-level priors (e.g., for \emph{atypical pigment network} in PH2, the Melanoma prior is $0.97$ vs.\ the prediction probability is $0.00$, indicating that the predictions do not line up with the expected distribution of concepts), while including $\mathcal{L}_{\text{KL}}$ aligns the prediction distribution more closely to the prior ($0.97$).  Similarly, removing $\mathcal{L}_{\text{ent}}$ yields higher entropy ($0.179$ for Melanoma), whereas applying $\mathcal{L}_{\text{ent}}$ reduces entropy to $0.009$, leading to sharper and more selective concept weighting. Both ablations lead to reduced accuracy and F1, while the full model with both regularizers achieves the best performance across datasets.

\begin{table}[h]
\centering
\caption{Classification performance (F1-score) across multiple datasets.}
\label{tab:class-results}
\scriptsize
\resizebox{\linewidth}{!}{
\begin{tabular}{lcccc}
\toprule
\textbf{Model} & \textbf{WBCatt (F1\%)} & \textbf{PH2 (F1\%)} & \textbf{HAM10000 (F1\%)} & \textbf{CXR4 (F1\%)} \\
\midrule
\rowcolor{gray!15}
BlackBox (ResNet) & $98.91_{\pm 0.32}$ & $95.00_{\pm 0.58}$ & $85.91_{\pm 0.61}$ & $88.23_{\pm 0.49}$ \\
Vanilla-CBM~\cite{koh2020concept} & $96.11_{\pm 0.24}$ & $77.14_{\pm 0.02}$ & - & - \\
Vanilla-V-IP~\cite{chattopadhyayvariational} & $96.31_{\pm 0.12}$ & $90.00_{\pm 0.78}$ & - & - \\
\hline
PCP-CBM (ours) & $95.23_{\pm 0.98}$ & $45.21_{\pm 0.25}$ & $77.74_{\pm 0.83}$ & $82.65_{\pm 0.41}$ \\
PCP-V-IP (ours) & $96.22_{\pm 1.01}$ & $87.50_{\pm 0.94}$ & $79.85_{\pm 0.55}$ & $84.32_{\pm 0.62}$ \\
\bottomrule
\end{tabular}}
\end{table}

\noindent \textbf{Classification Performance.} 
Based on Table~\ref{tab:class-results}, we observe that PCP-V-IP achieves classification performance comparable to Vanilla-V-IP across both PH2 and WBCatt. 
In contrast, PCP-CBM underperforms on PH2, primarily due to the small dataset size and noisy priors for key melanoma-related concepts such as \emph{atypical pigment network} (concept F1: 61.11\%) and \emph{irregular dots and globules} (concept F1: 47.61\%). 
Since CBM relies on all predicted concepts for classification, errors in these discriminative concepts directly degrade melanoma recognition. 
By contrast, V-IP naturally complements our prior-guided framework, as its query mechanism selectively focuses on the most reliable concepts and can bypass those that are poorly predicted from priors. 

For HAM10000 and CXR4, which lack concept annotations (required by Vanilla-CBM/V-IP), we compare against a ResNet black-box baseline.
Both PCP-CBM and PCP-V-IP achieve competitive F1-scores, indicating that meaningful concept representations can be learned directly from class-level priors without explicit concept supervision.
\begin{figure}[ht]
    \centering
    \includegraphics[width=\linewidth]{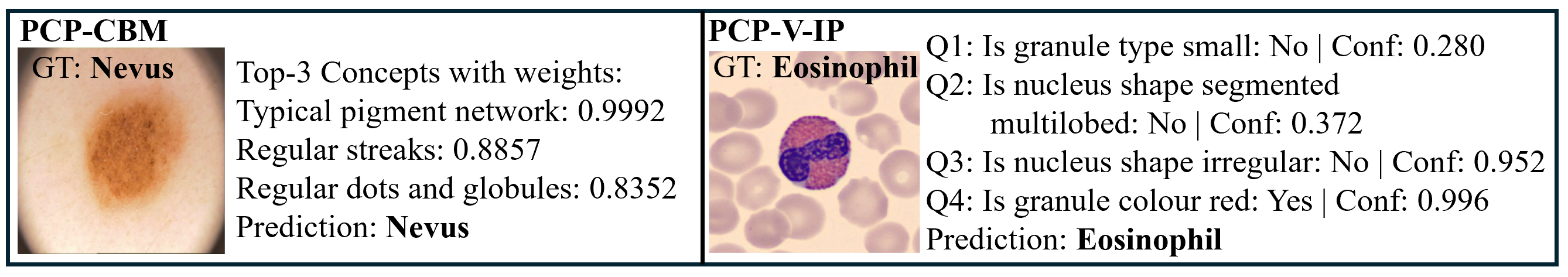}
    \caption{Qualitative interpretability examples using PCP.
    Left: CBM-based prediction on PH2.
    Right: V-IP-based prediction on WBCatt.}
    \label{fig:xai-examples}
\end{figure}

\noindent \textbf{Qualitative Analysis.}  
Fig.~\ref{fig:xai-examples} illustrates qualitative interpretability results on dermoscopy (PH2) and hematology (WBCatt). 
The model predicts clinically relevant concepts with high confidence, producing interpretable reasoning patterns consistent with expert knowledge, despite being trained without concept-level annotations.

% \vspace{0.5em}
\section{Conclusion} 
\label{sec:conclusion}
In this work, we present \textit{PCP}, a weakly supervised framework that enables medical concept prediction without explicit concept annotations or reliance on vision--language supervision. By leveraging class-level concept priors with KL and entropy regularization, PCP produces interpretable and clinically aligned concept predictions while maintaining competitive classification performance across multiple modalities. Although PCP relies on class-level priors, these are substantially easier to obtain than exhaustive concept annotations and can be derived from expert knowledge or dataset statistics, making the approach practical and scalable. In rare-disease or limited-data settings, prior quality may be noisier, which we identify as a key limitation and motivation for future work on adaptive prior refinement and self-distilled concept reasoning to improve robustness and generalization.

\bibliographystyle{unsrt}  
\bibliography{references}  %%% Remove comment to use the external .bib file (using bibtex).

\end{document}